\makeatletter\def\graphicscache@inhibit{true}\makeatother
\documentclass[conference]{IEEEtran}
\IEEEoverridecommandlockouts
\usepackage{cite}
\usepackage{amsmath,amssymb,amsfonts}
\usepackage{algorithmic}
\usepackage{graphicx}
\usepackage[dpi=300,qfactor=0.5]{graphicscache}
\usepackage{textcomp}
\usepackage{xcolor}

\usepackage{makeidx}

\usepackage{pgfplots}
\pgfplotsset{compat=newest}
\usepackage{multirow}
\usepackage{tabularx}
\usepackage{threeparttable}
\usepackage{booktabs}
\usepackage{array}
\usepackage{float}
\usepackage{graphbox}
\usepackage{tikz,tikz-3dplot}
\usetikzlibrary{fit,arrows,arrows.meta,automata,backgrounds,calc,chains,%
decorations.markings,decorations.pathreplacing,decorations.pathmorphing,%
matrix,positioning,shapes,shapes.geometric,shapes.symbols,spy,trees,tikzmark}
\usepackage{balance}
\usepackage[binary-units=true,product-units=single,per-mode=symbol,range-units=single,range-phrase=\,--\,,detect-all]{siunitx}
\DeclareSIUnit\pixel{px}
\usepackage{bm}
\usepackage{acronym}
\usepackage{tablefootnote}
\usepackage{hyperref}

\definecolor{bg_color}{RGB}{95,95,95}

\newcommand{\norm}[1]{\left\lVert#1\right\rVert}

\DeclareMathOperator*{\argmin}{arg\,min}

\newcommand{\reffig}[1]{Fig.~\ref{#1}}
\newcommand{\reftab}[1]{Tab.~\ref{#1}}
\newcommand{\refsec}[1]{Sec.~\ref{#1}}
\newcommand{\refeq}[1]{Eq.~\ref{#1}}

\usepackage{adjustbox}
\newcolumntype{R}[2]{%
    >{\adjustbox{angle=#1,lap=\width-(#2)}\bgroup}%
    l%
    <{\egroup}%
}
\newcolumntype{L}[1]{>{\raggedright\let\newline\\\arraybackslash\hspace{0pt}}m{#1}}

\begin{document}

\title{Object-level 3D Semantic Mapping using a Network of Smart Edge Sensors
\thanks{\hspace{-2.2ex}$^{*}$: equal contribution.}%
\thanks{This work was funded by grant BE 2556/16-2 (Research Unit FOR 2535 Anticipating Human Behavior) of the German Research Foundation (DFG).}
}

\author{\IEEEauthorblockN{Julian Hau$^{*}$}
\IEEEauthorblockA{\textit{Autonomous Intelligent Systems} \\
\textit{University of Bonn}\\
Bonn, Germany \\
Email: {\tt s6juhauu@uni-bonn.de}}
\and
\IEEEauthorblockN{Simon Bultmann$^{*}$}
\IEEEauthorblockA{\textit{Autonomous Intelligent Systems} \\
\textit{University of Bonn}\\
Bonn, Germany \\
Email: {\tt bultmann@ais.uni-bonn.de}}
\and
\IEEEauthorblockN{Sven Behnke}
\IEEEauthorblockA{\textit{Autonomous Intelligent Systems} \\
\textit{University of Bonn}\\
Bonn, Germany \\
Email: {\tt behnke@cs.uni-bonn.de}}
}

\maketitle

\begin{tikzpicture}[remember picture,overlay]
\node[anchor=north west,align=left,font=\sffamily,yshift=-0.2cm,xshift=0.2cm] at (current page.north west) {%
  6th IEEE International Conference on Robotic Computing (IRC), Naples, Italy, December 2022.
};
\end{tikzpicture}%
\vspace{-0.2cm}%

\begin{abstract}
Autonomous robots that interact with their environment require a detailed semantic scene model. For this, volumetric semantic maps are frequently used.
The scene understanding can further be improved by including object-level information in the map.
In this work, we extend a multi-view 3D semantic mapping system consisting of a network of distributed smart edge sensors with object-level information, to enable downstream tasks that need object-level input.
Objects are represented in the map via their 3D mesh model or as an object-centric volumetric sub-map that can model arbitrary object geometry when no detailed 3D model is available.
We propose a keypoint-based approach to estimate object poses via PnP and refinement via ICP alignment of the 3D object model with the observed point cloud segments. Object instances are tracked to integrate observations over time and to be robust against temporary occlusions. 
Our method is evaluated on the public Behave dataset where it shows pose estimation accuracy within a few centimeters and in real-world experiments with the sensor network in a challenging lab environment where multiple chairs and a table are tracked through the scene online, in real time even under high occlusions.
\end{abstract}


\section{Introduction}
\label{sec:Introduction}
Semantic scene understanding is an important prerequisite for many autonomous robotic tasks, like object manipulation or collision-free navigation.
For this, the environment is captured with various sensors and a semantic map is created from the interpreted measurements.
In this work, we extend a system for 3D semantic scene perception from prior work~\cite{bultmann_ias2022}, consisting of distributed smart edge sensors, with object-level information.
While the scene model from the previous work only comprises an allocentric semantic map without any information about object instances, we now represent objects in the map via their 3D mesh model. If no detailed 3D model is available, we create an object-centric volumetric sub-map that can model arbitrary object geometry. 
For this, we fuse the semantic percepts of four smart edge sensors with RGB-D cameras.
The sensor measurements are interpreted locally, on-device, by deep convolutional neural networks (CNNs) and the semantic object-level information is streamed to a central backend, where object detections from different views are fused and tracked through an allocentric scene model.
We employ a keypoint-based approach for object pose estimation~\cite{zappel20216d} using CNNs for keypoint detection trained only on synthetic data obtained through randomized scene generation~\cite{stilleben_2020,sl-cutscenes}. Object poses are recovered from keypoint detections via a variant of the PnP algorithm~\cite{EPnP} in each camera view and fused on the backend via weighted interpolation.
We evaluate our method on the public Behave dataset~\cite{behave_2022}, containing various scenes with human-object interactions, and in real-world experiments with the sensor network in a challenging, highly cluttered and dynamic lab environment.

\begin{figure}[t]
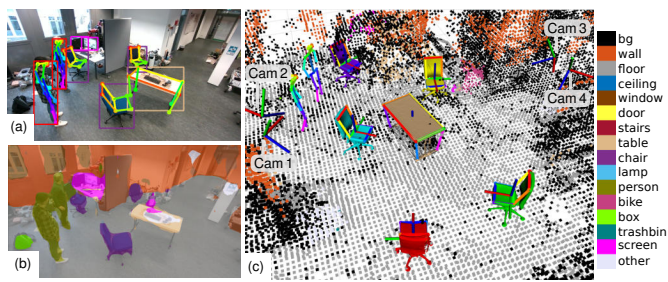

\centering
\begin{tikzpicture}
		    \node[inner sep=0,anchor=north west] (image1) at (0, 0){\includegraphics[height=3.6cm,trim={0 .5cm 0cm 0},clip]{img/teaser_new_3D_1.png}};
		    
		    \node[label, scale=.75, anchor=south west, xshift=0.1cm, yshift=-2.9cm, rectangle, rounded corners=2, inner sep=0.06cm, fill={white!90!black},opacity=.9,text opacity=1, align=center, font=\scriptsize\sffamily] (l_cam1) at (image1.north west) {Cam\,1};
			\draw[{white!90!black},opacity=.9, very thick] (l_cam1.140) ++(0., -0.006) -- ++(0., 0.2);
			
    		\node[label, scale=.75, anchor=south west, xshift=0.0cm, yshift=-1.3cm, rectangle, rounded corners=2, inner sep=0.06cm, fill={white!90!black},opacity=.9,text opacity=1, align=center, font=\scriptsize\sffamily] (l_cam2) at (image1.north west) {Cam\,2};
    		\draw[{white!90!black},opacity=.9, very thick] (l_cam2.220) ++(0., +0.006) -- ++(0., -0.35);
    		
    		\node[label, scale=.75, anchor=south west, xshift=5.3cm, yshift=-0.5cm, rectangle, rounded corners=2, inner sep=0.06cm, fill={white!90!black},opacity=.9,text opacity=1, align=center, font=\scriptsize\sffamily] (l_cam2) at (image1.north west) {Cam\,3};
    		\draw[{white!90!black},opacity=.9, very thick] (l_cam2.220) ++(0., 0.006) -- ++(0., -0.22);
    		
    		\node[label, scale=.75, anchor=south west, xshift=5.3cm, yshift=-1.75cm, rectangle, rounded corners=2, inner sep=0.06cm, fill={white!90!black},opacity=.9,text opacity=1, align=center, font=\scriptsize\sffamily] (l_cam2) at (image1.north west) {Cam\,4};
    		\draw[{white!90!black},opacity=.9, very thick] (l_cam2.70) ++(0., -0.006) -- ++(-0., 0.22);

			\node[inner sep=0,anchor=north west,xshift=0.05cm,yshift=-0.3cm] (image2) at (image1.north east) {\includegraphics[height=3.15cm]{img/legend_ade20k_indoor_reduced.pdf}};
			
			\node[inner sep=0,anchor=north east,xshift=-0.05cm] (image3) at (image1.north west) {\includegraphics[height=1.75cm]{img/teaser_new_2D_1.png}};
			\node[inner sep=0,anchor=north east,yshift=-0.05cm] (image4) at (image3.south east) {\includegraphics[height=1.75cm]{img/teaser_new_2Dsegm_1.png}};

			\node[label,scale=.75, anchor=south west,xshift=-0.07cm, rectangle, fill=white, align=center, font=\scriptsize\sffamily] (n_0) at (image1.south west) {(c)};
			\node[label,scale=.75, anchor=south west,xshift=-0.07cm, rectangle, fill=white, align=center, font=\scriptsize\sffamily] (n_1) at (image3.south west) {(a)};
			\node[label,scale=.75, anchor=south west, rectangle, fill=white, align=center, font=\scriptsize\sffamily] (n_3) at (image4.south west) {(b)};
\end{tikzpicture}
\vspace{-1.7em}
\caption{Object-level 3D semantic mapping: (a) object detection and keypoint estimation for the \textit{chair} and \textit{table} class; (b) semantic segmentation from an exemplary smart edge sensor view (Cam~2); (c) 3D scene view with five chairs and a table, represented by their resp. 3D mesh and colored by instance ID, in the allocentric semantic map~\cite{bultmann_ias2022} together with human keypoint poses~\cite{Bultmann_RSS_2021}.}
\label{fig:teaser}
\vspace{-1.2em}
\end{figure}

\reffig{fig:teaser} illustrates our approach, showing object detections, keypoint estimation for the \textit{chair} and \textit{table} class, and semantic segmentation from an exemplary smart edge sensor view together with the fused 3D semantic scene model with five chairs and a table represented by their respective 3D mesh tracked in the allocentric semantic map. To summarize, the main contributions of this paper are:
\begin{itemize}
\item A novel object-level 3D semantic mapping approach, fusing semantic information from multiple smart cameras,
\item a keypoint-based object pose estimation approach trained solely on synthetic data, and
\item quantitative evaluation of the pose estimation and geometry representation accuracy on the public Behave dataset and
 real-world experiments in a highly cluttered, dynamic lab environment.
\end{itemize}

\section{Related Work}
\label{sec:Related_Work}
\paragraph*{Semantic Mapping}
Different approaches exist in the literature to create three-dimensional maps to be used for localization and navigation of mobile robots.
A common approach are occupancy grid maps. Octomap~\cite{hornung_octomap_2013}, a widely-used 3D occupancy mapping framework, uses an efficient octree-based data structure to save occupancy probabilities of the environment divided into discrete volume elements (\textit{voxels}).
A second popular map representation are truncated signed distance functions (TSDFs). A TSDF map saves the distance to the closest surface in each voxel.
Voxblox~\cite{voxblox} is a commonly-used framework for building TSDF-based maps.

The above works, however, represent only geometry and don't contain any semantic information about the environment.
To extend geometric 3D maps with semantic information, Stückler et al.~\cite{stuckler_semantic_2012} fuse probabilistic segmentations from multiple RGB-D camera perspectives into a voxel-based 3D Map.
MaskFusion~\cite{maskFusion} is an RGB-D simultaneous localization and mapping (SLAM) system that can reconstruct and track multiple objects in a scene without knowing prior models of the objects. The constructed map uses surface elements (Surfels) to represent surfaces and Mask-RCNN~\cite{he_maskrcnn_2017} is used to obtain semantic instance segmentation of the RGB images.
MID-Fusion~\cite{mid} uses an octree-based TSDF-map to implement RGB-D SLAM. On top of the scene geometry, RGB-color, semantic classes, and a foreground probability are represented in the map.
Voxblox++~\cite{voxblox++} uses both geometric and semantic instance segmentation to build a semantic map with object-level information for static scenes.
With TSDF++~\cite{tsdf++}, Grinvald et al. proposed to create TSDF sub-volumes for object instances. The object sub-volumes are included in an allocentric volumetric map that can reference multiple objects at each location. Thus, temporally occluded objects or surfaces remain in the map and do not need to be reconstructed anew when being visible again and dynamic scenes can be represented.
The above works use a single, moving camera while we propose to fuse percepts from multiple static viewpoints.

Recently, Bultmann and Behnke~\cite{bultmann_ias2022} proposed an approach for fusing semantic segmentations from multiple RGB-D smart edge cameras to build an allocentric 3D semantic map. Together with the semantic map, 3D human poses are estimated in real time~\cite{Bultmann_RSS_2021} and represented in the scene model.
The camera images are processed locally, on the smart edge sensor boards, using embedded deep learning inference accelerators. Only semantic information is streamed over a network to a central backend. The raw images remain on the sensor board, significantly reducing the required communication bandwidth and providing scalability to a large number of sensor nodes.
The movement of dynamic objects is accounted for in the scene model via ray-tracing, enabling to remove freed voxels.
The ray-tracing update, however, reacts gradually and slowly to object movement and does not implement any instance-level object representation. Hence, object trajectories in the scene cannot be reconstructed.

In this work, we propose to represent objects in the map by their 3D mesh model or as an object-centric volumetric sub-map and robustly track their movement over time.

\paragraph*{Keypoint-based Pose Estimation}
For keypoint-based pose estimation, first, distinct points on the respective object model must be defined.
These keypoints are then detected in the camera image and the object pose is recovered via a variant of the PnP algorithm~\cite{EPnP}.
The assignment of detected keypoints to object instances is commonly implemented in two different manners: bottom-up or top-down.
Bottom-up approaches first detect keypoints and then assign them to object instances~\cite{cao_openpose_2018}.
Top-down approaches, on the other hand, first employ an object detector and then estimate keypoints on the image crop of each object~\cite{xiao_simple_2018}.
While the top-down approaches have the risk of early commitment due to errors in object detection and scale badly with a higher number of detected objects, bottom-up approaches can wrongly associate keypoints of different object instances and have difficulties handling objects of small scales.
Often, top-down approaches show higher accuracy but slower speed than bottom-up approaches~\cite{cao_openpose_2018}.
Zappel et al.~\cite{zappel20216d} used the bottom-up OpenPose approach~\cite{cao_openpose_2018} for keypoint detection on objects of the YCB-V dataset~\cite{PoseCNN}.
The 6\,DoF object poses were recovered using the PnP-RANSAC Algorithm~\cite{EPnP,Ransac} to calculate the object's translation and rotation in the camera frame from 2D--3D correspondences of keypoints on rigid objects.
Bultmann and Behnke~\cite{Bultmann_RSS_2021,bultmann_ias2022} used a top-down approach for human pose estimation on embedded smart edge sensors employing the CNN architecture of Xiao et al.~\cite{xiao_simple_2018} with an efficient MobileNet\,V3~\cite{mobilenetv32019} backbone. 3D human poses are recovered from 2D keypoint detections via multi-view triangulation.
In this work, we adopt the top-down approach for object keypoint detection on embedded smart edge sensors and implement object pose estimation via the PnP-RANSAC Algorithm. Pose estimates from multiple sensor views are fused via weighted interpolation.

\section{Method}
\label{sec:Method}
\begin{figure*}[t]
	\centering  
        	\includegraphics[width=\textwidth,trim={0.5cm 13.7cm 3.5cm 1cm},clip]{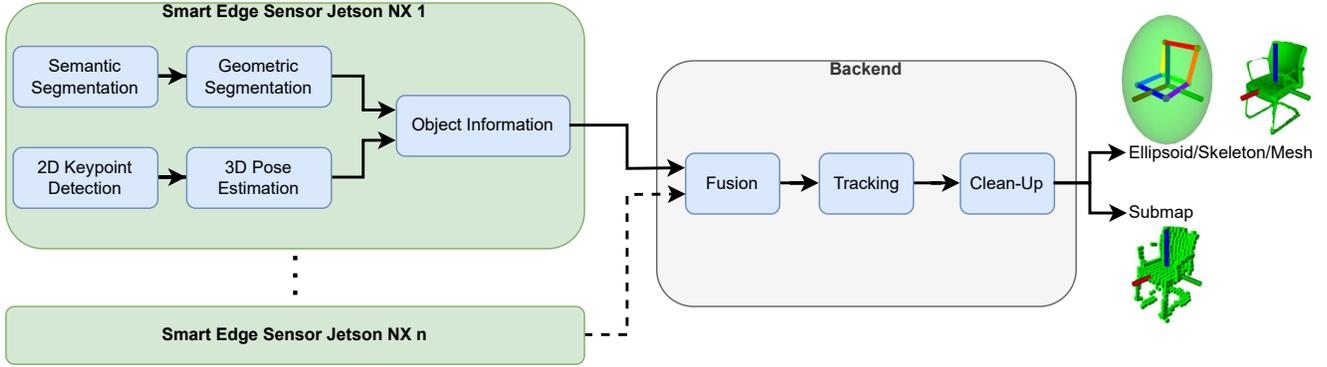}
        \vspace{-2.em}
       	\caption{Overview of the object-level mapping pipeline: Smart edge sensors generate semantically and geometrically segmented point clouds. Simultaneously, 3D object poses are estimated via PnP using 2D keypoint detections. With the results, we calculate information about each observed instance. On the backend, observations from multiple views are fused and objects are tracked over time.}
        \label{fig:pipeline}
        
\end{figure*}
Our method for object-level semantic mapping uses multi-view data from $N$ smart edge sensors with RGB-D cameras. The sensor hardware comprises an Nvidia Jetson NX embedded compute board and an Intel RealSense D455 RGB-D camera, as introduced in~\cite{bultmann_ias2022}. The camera poses in world coordinates are assumed to be calibrated beforehand~\cite{paetzold_camcalib_2022}.
\reffig{fig:pipeline} gives an overview of our proposed pipeline for object-level mapping.
In each sensor view, point clouds for the considered object classes are extracted via semantic segmentation of the RGB image and projection of the depth data~\cite{bultmann_ias2022}. The object point clouds are then geometrically segmented via the Euclidean cluster algorithm and a statistical outlier filter~\cite{pcl_icra_2011} to obtain a point cloud segment per detected object instance.
Simultaneously, object keypoint detection is performed on the RGB images and 6\,DoF object poses are recovered via the PnP-RANSAC Algorithm~\cite{zappel20216d} using 2D--3D correspondences between detected image keypoints and 3D object model keypoints. We assume 3D models of the considered object classes to be available as prior information on the sensor boards and the backend.
The keypoint-based pose estimates are then associated to point cloud segments via the closest distance between object keypoints and their nearest neighbor in the respective point cloud segment. The pose estimate is further refined via ICP alignment of the object model, initialized with the PnP pose estimate, with the corresponding point cloud segment.
Various semantic object properties are calculated from the associated point cloud segments and keypoint detections. 

The semantic object information is streamed to a central backend, where the observations from multiple smart edge sensors are fused.
We discern two different cases, depending on the available network bandwidth and the used object representation in the scene model:
The transmitted object information always comprises the pose estimate, the mean distance between object keypoints and nearest neighbors in the associated point cloud segment, and the statistical distribution of the points in the cluster, visualized as a covariance ellipsoid anchored at the object model origin.
Objects are represented in this case by a 3D keypoint skeleton with an associated point distribution ellipsoid or by their 3D mesh model.

If a 3D mesh model is not available, objects are represented by an object-centric volumetric sub-map, which can model arbitrary geometric shapes. For this, the point cloud segments are additionally transmitted to the backend, requiring higher network bandwidth (cf. \refsec{sec:eval_datarate}).

A tracking module enables to robustly follow object trajectories through the scene and a clean-up step removes object hypotheses that moved out of view or were falsely initialized from noisy measurements.

\subsection{Keypoint-based Object Pose Estimation}
\label{sec:pose_est}
\begin{figure}[t]
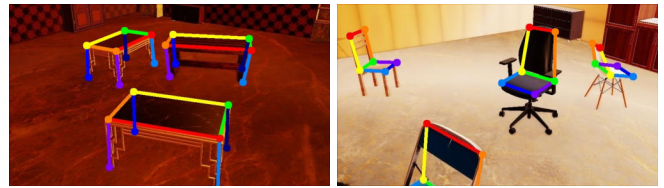

	\centering
	\begin{tikzpicture}
		\node[inner sep=0,anchor=north west] (image1) at (0, 0){\includegraphics[width = 0.48\linewidth]{img/scene1_table.jpg}};
		\node[inner sep=0,anchor=north west,xshift=0.1cm] (image2) at (image1.north east) {\includegraphics[width = 0.48\linewidth]{img/scene2.jpg}};
	\end{tikzpicture}
	\vspace{-.5em}
        \caption{Frames from two training scenes with corresponding ground truth keypoint annotations for the \textit{table} or \textit{chair} class, respectively. Background textures and object models are randomly selected.}
        \label{fig:trainingscenes}
\end{figure}

For keypoint-based pose estimation, keypoint locations need to be defined at distinct points of the object model.
We perform keypoint estimation for the \textit{chair} and \textit{table} object classes in this work and aim to represent different types of chairs and tables with the defined keypoints.
For this, we define $L_\text{chair}=6$ keypoints on chairs: four keypoints on the corners of the seating and two keypoints at the top of the backrest of a chair. These keypoints can be consistently defined for most types of chairs and have less variance in appearance and geometry than, e.g., armrests or legs of chairs.
Similarly, we define $L_\text{table}=8$ keypoints on tables: four keypoints on the corners of the tabletop and four on the points of the table legs (cf. \reffig{fig:trainingscenes}).

We train a deep neural network for keypoint detection on RGB images for chairs and tables, respectively, using the network architecture of Xiao et al.~\cite{xiao_simple_2018} with a MobileNet\,V3 backbone~\cite{mobilenetv32019} that proved efficient on embedded hardware in prior works~\cite{Bultmann_RSS_2021,bultmann_ias2022}.
As we follow a top-down approach for keypoint estimation, an object detector is required prior to the keypoint estimation step, to provide bounding boxes used to extract the input object crops. For this, we employ a MobileDet detector~\cite{xiong_mobiledets_2021} trained on the COCO dataset~\cite{lin_coco_2014}, as in~\cite{bultmann_ias2022}.

To avoid costly manual annotation of training data and to facilitate generalization to different object classes, we only use synthetic training images for the keypoint estimation. We employ the \textit{sl-cutscenes} framework~\cite{sl-cutscenes}, an extension of the \textit{stillleben} framework~\cite{stilleben_2020}, for randomized photorealistic indoor scene generation with physically interacting objects. We create a dataset of $\sim$13k training and $\sim$2.5k validation images per object class where between three and six randomly selected chairs or tables move around a room with randomly selected textures and background objects. \reffig{fig:trainingscenes} shows samples of the generated training images.
\begin{figure*}[t]
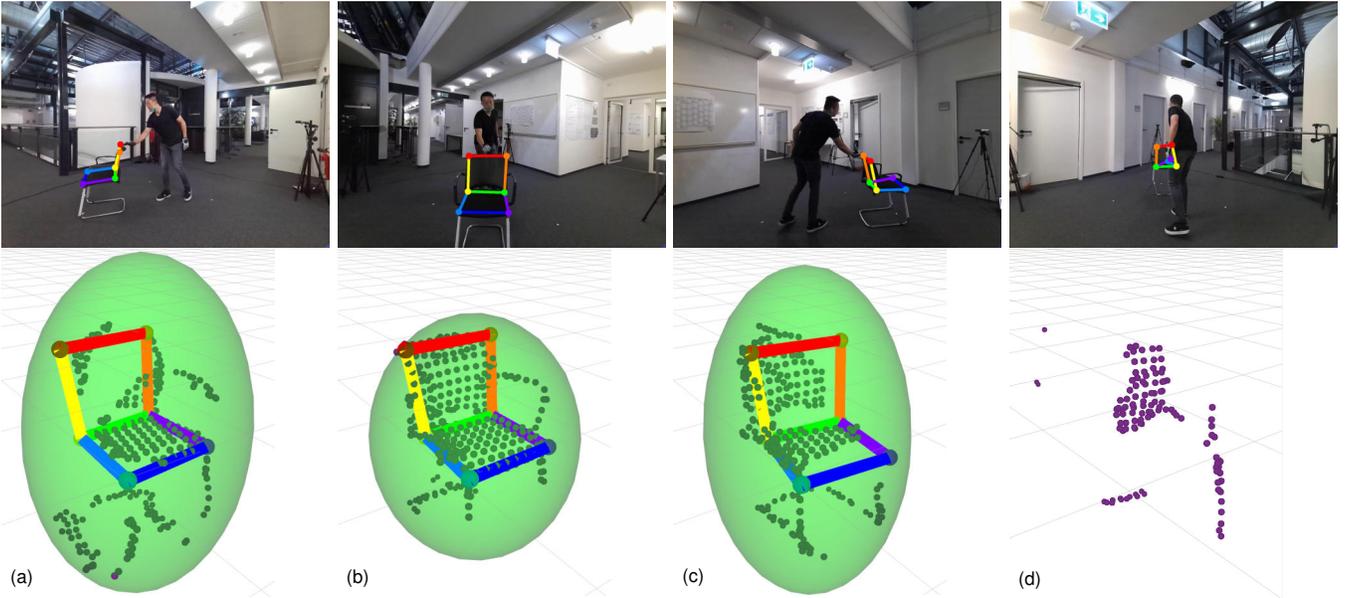
 
	\centering
	\begin{tikzpicture}
		\node[inner sep=0,anchor=north west] (image1) at (0, 0){\includegraphics[width = 0.24\textwidth]{img/2d_keypoints1.jpg}};
		\node[inner sep=0,anchor=north west,xshift=0.1cm] (image2) at (image1.north east) {\includegraphics[width = 0.24\textwidth]{img/2d_keypoints2.jpg}};
		\node[inner sep=0,anchor=north west,xshift=0.1cm] (image3) at (image2.north east) {\includegraphics[width = 0.24\textwidth]{img/2d_keypoints3.jpg}};
		\node[inner sep=0,anchor=north west,xshift=0.1cm] (image4) at (image3.north east) {\includegraphics[width = 0.24\textwidth]{img/2d_keypoints4.jpg}};
		
		\node[inner sep=0,anchor=north west,yshift=-0.01cm] (image5) at (image1.south west) {\includegraphics[width = 0.2\textwidth,trim={0 1.5cm 0 0.5cm},clip]{img/keypoints_aligned_1.png}};
		\node[inner sep=0,anchor=north west,yshift=-0.01cm] (image6) at (image2.south west) {\includegraphics[width = 0.2\textwidth,trim={0 1.5cm 0 0.5cm},clip]{img/keypoints_aligned_2.png}};
		\node[inner sep=0,anchor=north west,yshift=-0.01cm] (image7) at (image3.south west) {\includegraphics[width = 0.2\textwidth,trim={0 1.5cm 0 0.5cm},clip]{img/keypoints_aligned_3.png}};
		\node[inner sep=0,anchor=north west,yshift=-0.01cm] (image8) at (image4.south west) {\includegraphics[width = 0.2\textwidth,trim={0 3.cm 0 1.cm},clip]{img/pnp_keypoints4.png}};
		
		\node[label,scale=1., anchor=south west, rectangle, fill=white, align=center, font=\scriptsize\sffamily] (n_0) at (image5.south west) {(a)};
		\node[label,scale=1., anchor=south west, rectangle, fill=white, align=center, font=\scriptsize\sffamily] (n_1) at (image6.south west) {(b)};
		\node[label,scale=1., anchor=south west, rectangle, fill=white, align=center, font=\scriptsize\sffamily] (n_3) at (image7.south west) {(c)};
		\node[label,scale=1., anchor=south west, rectangle, fill=white, align=center, font=\scriptsize\sffamily] (n_3) at (image8.south west) {(d)};
	\end{tikzpicture}
	\vspace{-.5em}
	\caption{2D keypoint detections and corresponding 3D poses calculated with PnP-RANSAC and ICP refinement. Top-row: keypoint detections in four different perspectives  of a scene from the Behave dataset~\cite{behave_2022}. Bottom-row: object pose estimation and associated point cloud segments. For (a)--(c), the method was able to estimate a pose for the object, while no valid pose estimate and data association could be obtained for perspective (d) due to the high occlusion.} 
        \label{fig:pnp-pose}
\end{figure*}

The detected object keypoints are used in a second step to recover the object's translation and rotation in the camera coordinates via the PnP-RANSAC algorithm, as proposed by Zappel et. al~\cite{zappel20216d}.
For this, we assume a 3D model of the specific type of chair or table visible in the respective scene to be available and exploit the correspondences between 2D image keypoint and keypoints defined on the 3D object model.
A 3D object skeleton in camera coordinates is then obtained by transforming the object model keypoints with the estimated PnP pose.
As we consider chairs and tables standing or moving on the ground plane, the PnP pose estimate is further projected to the ground plane ($xy$-plane in allocentric coordinates), to obtain a stable and plausible object pose, compensating for noise or outliers in the keypoint detections.

In parallel to the keypoint detection on RGB images, point cloud segments of the observed objects are obtained from the depth data via semantic and geometric segmentation (cf.~\reffig{fig:pipeline}).
To fuse the keypoint-based pose estimate with the point cloud observations, 3D object skeletons are assigned to point cloud segments in a data association step.
For each frame, we obtain a set of 3D keypoint skeletons $\mathcal{K}$ and a set of point cloud segments $\mathcal{S}$ of the corresponding semantic class.
For each segment $s_j\in\mathcal{S}$, we find the corresponding object skeleton $k_i\in\mathcal{K}$ with the minimum average distance between 3D object keypoints $x_{l,k_i}$ and their nearest neighbor in the respective point cloud segment $y_{l,s_j}$:
\begin{align}
d_\text{kps-segm}\left(k_i, s_j\right) = \frac{1}{L} \sum_{l=1}^{L} \norm{x_{l,k_i} - y_{l,s_j}}\,,\label{eqn:data_assoc_dist} \\
k_\text{min} = \argmin_{\forall k_i \in \mathcal{K}} \left( d_\text{kps-segm}\left(k_i, s_j\right) \right)\,. \label{eqn:data_assoc}
\end{align}

Data association is performed in a greedy manner, starting with the largest point cloud segment, and associations are valid only when the obtained average keypoint to nearest point cloud neighbor distance $d_\text{kps-segm}(k_\text{min}, s_j)$ is below a threshold $\tau_\text{dist}$.
Segments and keypoint detections without a valid data association are discarded.

Lastly, to further improve the estimated object pose, the keypoint-based PnP pose estimate is refined via ICP alignment with the point cloud data.
The object model point cloud, sampled from the 3D object mesh model and initialized with the PnP pose, is aligned with the associated observed point cloud segment, resulting in the final object pose estimate.

The semantic object information, comprising the refined pose estimate, the data association distance $d_\text{kps-segm}(k_\text{min}, s_j)$, and the statistical distribution of the points in the cluster, is then streamed to a central backend, where object observations from multiple sensor perspectives are synchronized and fused.
If the sub-map representation is chosen on the backend (cf. \refsec{sec:sem_mapping}), the associated point cloud segments are additionally transmitted to the backend, significantly increasing the used network bandwidth (cf. \refsec{sec:eval_datarate}).
\reffig{fig:pnp-pose} shows object keypoint detections and resulting pose estimates with associated point cloud segments for an exemplary scene of the Behave dataset~\cite{behave_2022} with the \textit{chair} object class.

\subsection{Multi-view Object-level Semantic Mapping}
\label{sec:sem_mapping}
\begin{figure}[t]
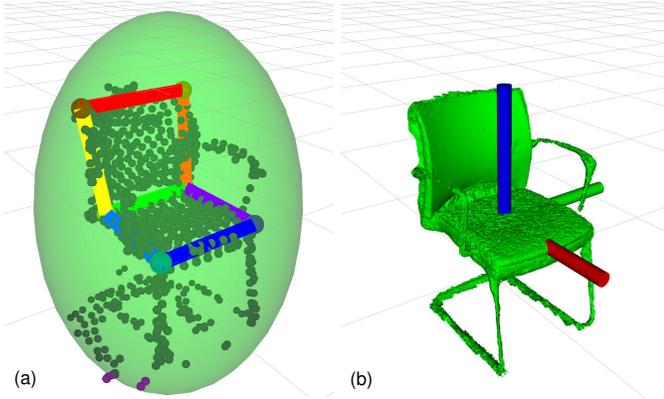

	\centering
	\begin{tikzpicture}
	\node[inner sep=0,anchor=north west] (image1) at (0, 0){\includegraphics[width = 0.24\textwidth,trim={0 2.5cm 0 1cm},clip]{img/fused_keypoints.png}};
	\node[inner sep=0,anchor=north west,xshift=0.1cm] (image2) at (image1.north east) {\includegraphics[width = 0.24\textwidth,trim={0 2.5cm 0 1cm},clip]{img/fused_mesh.png}};
	
	\node[label,scale=1., anchor=south west, rectangle, fill=white, align=center, font=\scriptsize\sffamily] (n_0) at (image1.south west) {(a)};
    \node[label,scale=1., anchor=south west, rectangle, fill=white, align=center, font=\scriptsize\sffamily] (n_1) at (image2.south west) {(b)};
	\end{tikzpicture}
	\vspace{-1.5em}
    \caption{Fusion of keypoint poses and point cloud variance: (a) fused keypoint skeleton and the point cloud variance from smart edge sensor observations (cf. \reffig{fig:pnp-pose}); (b) object mesh transformed with the fused pose estimate. Merged point cloud segments are shown as a reference in (a).} 
        \label{fig:fused-skeleton}
\end{figure}

A central backend receives semantic object pose and shape information from multiple smart edge sensor views.
The data streams are software-synchronized according to their time\-stamps.
Fused object pose estimates are obtained by i)~transforming the object pose estimates of individual cameras to allocentric coordinates, using the known extrinsic camera calibration, and ii)~weighted interpolation between the sensor views. The interpolation weights are inversely proportional to the data association distance (\refeq{eqn:data_assoc_dist}), giving the highest confidence to perspectives where the keypoint-based pose estimate is most consistent with the point cloud segments.
Spherical linear interpolation of quaternions is used for the orientations.
The point segment distribution variance parameters are averaged using the same interpolation weights.
Observations from at least one sensor view are required for a valid object instance. The fusion of multiple perspectives increases the robustness and accuracy of the pose and shape estimation.
\reffig{fig:fused-skeleton} shows the fused pose estimate using the individual poses from the three valid perspectives of \reffig{fig:pnp-pose}.
\begin{figure}[t]
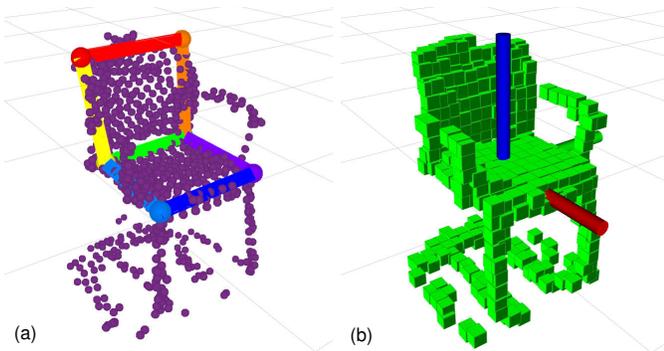

	\centering
	\begin{tikzpicture}
	\node[inner sep=0,anchor=north west] (image1) at (0, 0){\includegraphics[width = 0.24\textwidth,trim={0 2cm 0 5cm},clip]{img/fused_pointcloud.png}};
	\node[inner sep=0,anchor=north west,xshift=0.1cm] (image2) at (image1.north east) {\includegraphics[width = 0.24\textwidth,trim={0 2cm 0 5cm},clip]{img/fused_submap.png}};
	
	\node[label,scale=1., anchor=south west, rectangle, fill=white, align=center, font=\scriptsize\sffamily] (n_0) at (image1.south west) {(a)};
    \node[label,scale=1., anchor=south west, rectangle, fill=white, align=center, font=\scriptsize\sffamily] (n_1) at (image2.south west) {(b)};
	\end{tikzpicture}
	\vspace{-1.5em}
    \caption {Sub-map update with fused point cloud data: The point measurements of the merged point cluster (a) are integrated into a volumetric sub-map (b). Simultaneously, the sub-map is updated with the estimated object pose.} 
    \label{fig:fused-submap}
    \vspace{-1.em}
\end{figure}

\begin{figure}[t]
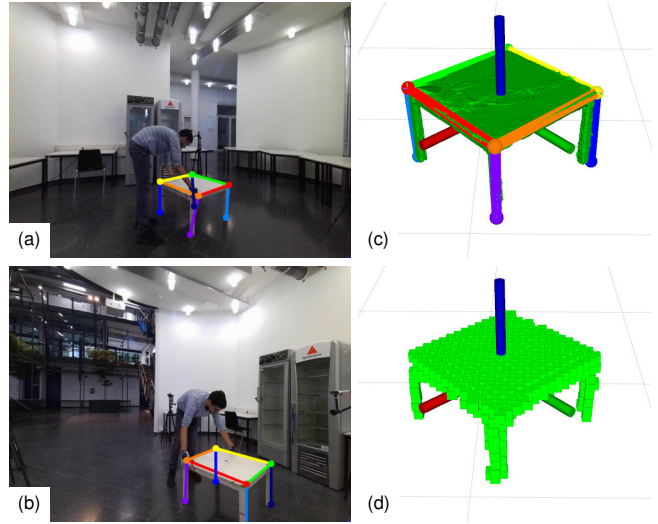

	\centering
	\begin{tikzpicture}
	\node[inner sep=0,anchor=north west] (image1) at (0, 0){\includegraphics[height = 3.4cm,trim={0 0cm 0 0cm},clip]{img/t0025_000_k0_color.jpg}};
	\node[inner sep=0,anchor=north west,yshift=-0.1cm] (image2) at (image1.south west) {\includegraphics[height = 3.4cm,trim={0 0cm 0 0cm},clip]{img/t0025_000_k1_color.jpg}};
	
	\node[inner sep=0,anchor=north west,xshift= 0.1cm] (image3) at (image1.north east) {\includegraphics[height = 3.4cm,trim={14cm 8cm 16cm 7cm},clip]{img/mesh_only.png}};
	\node[inner sep=0,anchor=north west,yshift= -0.1cm] (image4) at (image3.south west) {\includegraphics[height = 3.4cm,trim={14cm 8cm 16cm 7cm},clip]{img/submap_only.png}};
	
	\node[label,scale=1., anchor=south west, rectangle, fill=white, align=center, font=\scriptsize\sffamily] (n_0) at (image1.south west) {(a)};
    \node[label,scale=1., anchor=south west, rectangle, fill=white, align=center, font=\scriptsize\sffamily] (n_1) at (image2.south west) {(b)};
    \node[label,scale=1., anchor=south west, rectangle, fill=white, align=center, font=\scriptsize\sffamily] (n_2) at (image3.south west) {(c)};
    \node[label,scale=1., anchor=south west, rectangle, fill=white, align=center, font=\scriptsize\sffamily] (n_3) at (image4.south west) {(d)};
    
	\end{tikzpicture}
	\vspace{-.3em}
    \caption {2D keypoint detections for \textit{table} class from two perspectives of Behave dataset (a, b); fused 3D pose represented through (c) mesh or (d) submap.} 
    \label{fig:table-behave}
    \vspace{-.7em}
\end{figure}

If the sub-map representation is used for objects in the allocentric map, an object-centric sub-volume is maintained for each object instance, using a sparse voxel grid data structure, similar to the allocentric semantic map~\cite{bultmann_ias2022}.
Its size and resolution are chosen according to the represented object, independent of the resolution of the allocentric map. We use a voxel edge length of \SI{5}{cm} in our experiments.
To initialize and update the object sub-map, the point cloud segments associated to the detected object instances are additionally transmitted to the backend.
In a first step, point cloud segments from individual views are transformed to allocentric coordinates, using the camera extrinsics, and concatenated to form a merged object point cluster.
The merged object point cluster is then transformed into local object coordinates using the fused object pose estimate and integrated into the object-centric sub-map.
Each point measurement increments the occupancy count of its corresponding voxel. Once the occupancy is above a threshold $\tau_\text{occ}$, the voxel is considered occupied. The updated local sub-map is displayed at the estimated object pose in the allocentric scene model.
\reffig{fig:fused-submap} illustrates the sub-map update with the fused point cloud cluster from the three valid perspectives of \reffig{fig:pnp-pose}.
\reffig{fig:table-behave} shows 2D keypoint detections and fused 3D pose estimate for a sample scene of the Behave dataset with the \textit{table} object class.

Object instances are tracked over time on the backend via data association using a constant velocity model.
The position of known objects is predicted using a moving average velocity, computed over a fixed time window of past positions, and the passed time $\Delta t$ since the last synchronized frame-set was received from the sensors.
For each observed object instance, the nearest neighbor from the tracked object hypotheses is used, using their predicted position. Data associations are valid only if the distance between observation and corresponding track is below a threshold $\tau_\text{track}$.
For observations with no valid association, new tracking hypotheses are initialized.
In a clean-up step, object hypotheses that have not been observed for a longer time are removed. %

\section{Evaluation}
\label{sec:Evaluation}
We evaluate our approach on parts of the public Behave dataset~\cite{behave_2022}. The dataset comprises 321 video sequences totaling $\sim$15k frames, captured from four Kinect RGB-D cameras at a frame rate of \SI{1}{\hertz}. In the different scenarios, eight different persons interact with 20 different objects in five different environments. For each frame, annotations of the camera poses, 2D object and person segmentation masks, and pseudo-ground-truth object poses are available.

\subsection{Accuracy of Pose and Geometry Estimation}
We evaluate the accuracy of the proposed keypoint-based pose estimation with ICP refinement on five sequences of the Behave dataset, comprising $\sim$1k frames, containing interactions of a person with two different chair and one table models. We use the 3D mesh models of the chairs and table provided by the dataset as the basis for the PnP pose estimate.
\begin{table}[t]
\caption{Pose evaluation for scenarios with ground-truth segmen\-tation: Translation error (cm) and rotation error ($^\circ$).}
\vspace{-.5em}
\label{tab:gt-segmentation-pose}
\centering
\setlength{\tabcolsep}{4pt}
\begin{threeparttable}
\begin{tabular}{c|c|cc|cc}
  \toprule
  Scenario & Type & $E_\text{trans}$ & $\sigma_\text{trans}$ &  $E_\text{rot}$ & $\sigma_\text{rot}$ \\
  \midrule
  \multirow{3}{*}{\shortstack{\textit{chairblack}\\\textit{hand}}} & PnP only & 5.87 & 3.60 & 4.48 & 3.98 \\
  & PnP + ICP (local) & 3.40 & 3.08 & \textbf{3.33} & 2.78 \\
  & PnP + ICP (backend) & \textbf{2.88} & 3.11 & 3.52 & 2.92 \\\midrule

  \multirow{3}{*}{\shortstack{\textit{chairblack}\\\textit{sit}}} & PnP only & 5.23 & 2.46 & \textbf{5.64} & 6.97 \\
  & PnP + ICP (local) & \textbf{5.03} & 2.70 & 5.74 & 6.95 \\
  & PnP + ICP (backend) & 6.16 & 2.38 & 6.24 & 6.78 \\\midrule

  \multirow{3}{*}{\shortstack{\textit{chairwood}\\\textit{hand}}} & PnP only & 10.79 & 6.21 & 10.09 & 11.14 \\
  & PnP + ICP (local) & 6.37 & 4.68 & 8.23 & 11.37 \\
  & PnP + ICP (backend) & \textbf{6.35} & 4.73 & \textbf{6.16} & 8.73 \\\midrule

  \multirow{3}{*}{\shortstack{\textit{chairwood}\\\textit{sit}}} & PnP only & 13.17 & 5.03 & 7.14 & 6.23 \\
  & PnP + ICP (local) & 5.50 & 1.31 & 5.44 & 5.98 \\
  & PnP + ICP (backend) & \textbf{5.30} & 1.53 & \textbf{5.38} & 5.73 \\\midrule

  \multirow{3}{*}{\shortstack{\textit{tablesquare}\\\textit{move}}} & PnP only &  12.56 & 7.81 & 17.51 & 11.65   \\
  & PnP + ICP (local) & \textbf{7.82} & 8.29 & \textbf{3.37} & 3.69   \\
  & PnP + ICP (backend) & 8.01 & 8.19 & 3.93 & 6.08  \\
  
  \bottomrule
\end{tabular}
\end{threeparttable}
\end{table}
\begin{table}[t]
\caption{Pose evaluation for scenarios with online segmentation and detection: Translation error (cm) and rot. error ($^\circ$).}
\vspace{-.5em}
\label{tab:online-segmentation-pose}
\centering
\setlength{\tabcolsep}{4pt}
\begin{threeparttable}
\begin{tabular}{c|c|cc|cc}
  \toprule
  Scenario & Type & $E_\text{trans}$ & $\sigma_\text{trans}$ &  $E_\text{rot}$ & $\sigma_\text{rot}$ \\
  \midrule
  \multirow{3}{*}{\shortstack{\textit{chairblack}\\\textit{hand}}} & PnP only & 7.48 & 9.22 & 7.27 & 8.49 \\
  & PnP + ICP (local) & \textbf{6.99} & 8.23 & \textbf{6.50} & 10.05 \\
  & PnP + ICP (backend) & 7.42 & 7.93 & 6.75 & 9.53 \\\midrule

  \multirow{3}{*}{\shortstack{\textit{chairblack}\\\textit{sit}}} & PnP only & 8.09 & 6.49 & 11.34 & 13.02 \\
  & PnP + ICP (local) & \textbf{7.54} & 6.48 & \textbf{8.71} & 8.58 \\
  & PnP + ICP (backend) & 8.18 & 6.53 & 10.79 & 11.34 \\\midrule

  \multirow{3}{*}{\shortstack{\textit{chairwood}\\\textit{hand}}} & PnP only & 9.05 & 6.05 & 10.52 & 15.37 \\
  & PnP + ICP (local) & \textbf{6.47} & 4.42 & 10.51 & 15.16 \\
  & PnP + ICP (backend) & 7.36 & 4.04 & \textbf{8.32} & 14.09 \\\midrule

  \multirow{3}{*}{\shortstack{\textit{chairwood}\\\textit{sit}}} & PnP only & 5.42 & 3.40 & 8.52 & 4.63 \\
  & PnP + ICP (local) & \textbf{5.38} & 3.04 & 6.96 & 3.37 \\
  & PnP + ICP (backend) & 5.50 & 3.17 & \textbf{6.29} & 2.56 \\\midrule

  \multirow{3}{*}{\shortstack{\textit{tablesquare}\\\textit{move}}} & PnP only &  15.76 & 11.12 & 18.34 & 11.39  \\
  & PnP + ICP (local) &  \textbf{8.65} & 6.85 & \textbf{6.83} & 8.95   \\
  & PnP + ICP (backend) &  8.95 & 6.51 & 7.52 & 9.19  \\

  \bottomrule
\end{tabular}
\end{threeparttable}
\vspace{-.8em}
\end{table}
We calculate translation and orientation error w.r.t. to the pseudo-ground-truth object poses from the dataset, using the quaternion geodesic distance for the orientation, and compare the PnP-only raw pose estimate with the proposed ICP refinement in \reftab{tab:gt-segmentation-pose} and \reftab{tab:online-segmentation-pose}.
We discern two different options for the pose refinement via ICP alignment: refinement locally on the sensor boards, as described in \refsec{sec:pose_est}, and refinement with the merged point cluster on the backend. The latter requires transmitting the point cloud segments.

\reftab{tab:gt-segmentation-pose} reports the mean and standard deviation of the pose error when using the ground-truth point cloud segments and object boxes from the dataset as input to our keypoint estimation CNNs. The evaluation thus focuses on the keypoint-based pose estimation part, excluding other error sources present in real-world input data.
The ICP-based pose refinement significantly improves the translation error in all cases and the orientation error in all but one scenario. There are only little differences in accuracy between the refinement locally on the sensor board and on the backend.

\reftab{tab:online-segmentation-pose} reports the mean and standard deviation of the pose error when using online segmentation and object detection together with our keypoint estimation CNNs and thus evaluates the method's performance in real-world conditions.
The ICP-based refinement again significantly decreases both translation and orientation errors in all scenarios. The ICP-refinement locally on the sensor boards consistently performs better w.r.t. the translation error than refinement on the backend.
Therefore, we decide to use the local ICP refinement for further evaluation and real-world experiments.
Furthermore, in this case, the transmission of the point cloud segments is not necessary, when no volumetric sub-maps are needed, significantly decreasing the required network bandwidth (cf. \refsec{sec:eval_datarate}).
\begin{figure}[t]
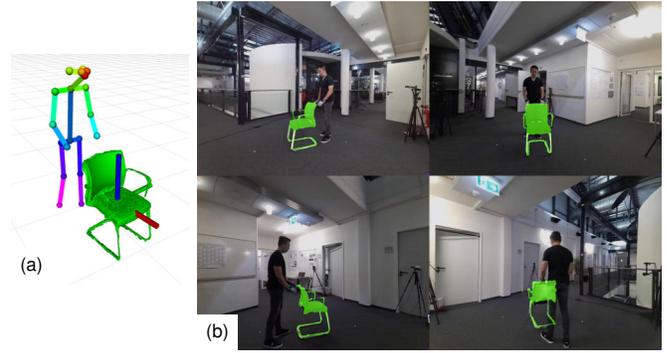

	\centering
	\begin{tikzpicture}
	\node[inner sep=0,anchor=north west] (image1) at (0, 0){\includegraphics[width = 0.13\textwidth]{img/backproj_mesh_3d.png}};
	\node[inner sep=0,anchor=north west,yshift=.7cm,xshift=0.1cm] (image2) at (image1.north east) {\includegraphics[width = 0.34\textwidth]{img/mesh_projected.png}};
	
	\node[label,scale=1., anchor=south west, rectangle, fill=white, align=center, font=\scriptsize\sffamily] (n_0) at (image1.south west) {(a)};
    \node[label,scale=1., anchor=south west, rectangle, fill=white, align=center, font=\scriptsize\sffamily] (n_1) at (image2.south west) {(b)};
	\end{tikzpicture}   
	\vspace{-.4em}
    \caption{Mesh backprojected into the camera images: The object model mesh (a) transformed with the estimated object pose is rendered in each camera view (b). The mesh is a 3D scan of the used object. Therefore, even fine elements align well.} 
        \label{fig:backprojection-mesh}
        
\end{figure}
\begin{figure}[t]
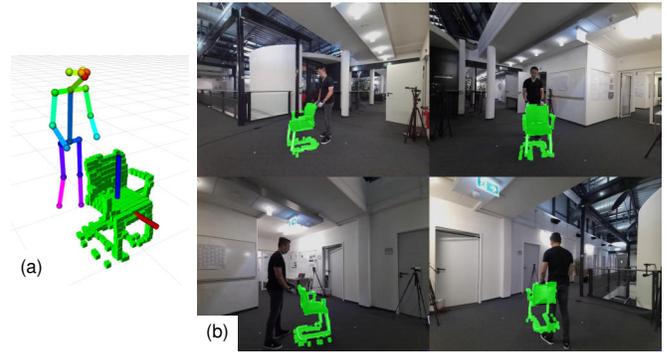

	\centering
	\begin{tikzpicture}
	\node[inner sep=0,anchor=north west] (image1) at (0, 0){\includegraphics[width = 0.13\textwidth]{img/backproj_submap_3d.png}};
	\node[inner sep=0,anchor=north west,yshift=.7cm,xshift=0.1cm] (image2) at (image1.north east) {\includegraphics[width = 0.34\textwidth]{img/submap_projected.png}};
	
	\node[label,scale=1., anchor=south west, rectangle, fill=white, align=center, font=\scriptsize\sffamily] (n_0) at (image1.south west) {(a)};
    \node[label,scale=1., anchor=south west, rectangle, fill=white, align=center, font=\scriptsize\sffamily] (n_1) at (image2.south west) {(b)};
	\end{tikzpicture}
	\vspace{-.4em}
    \caption{Sub-map backprojected into the camera images: Each voxel of the sub-map (a) is backprojected into the individual camera views (b). Because of the discrete resolution, fine elements like the legs are not accurately represented.} 
        \label{fig:backprojection-submap}
    \vspace{-1.em}
\end{figure}
\begin{figure*}[t]
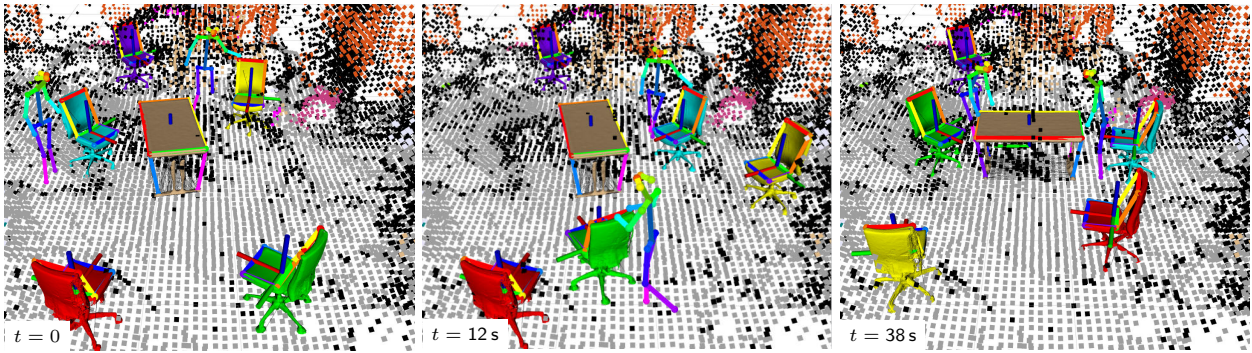

	\centering
	\begin{tikzpicture}
	\node[inner sep=0,anchor=north west] (image1) at (0, 0){\includegraphics[width = 0.3\textwidth,trim={1.5cm 1.5cm 0cm 1.cm},clip]{img/move_mesh_table_1.png}};
	\node[inner sep=0,anchor=north west,xshift=0.1cm] (image2) at (image1.north east) {\includegraphics[width = 0.3\textwidth,trim={1.5cm 1.5cm 0cm 1.cm},clip]{img/move_mesh_table_2_1.png}};
	\node[inner sep=0,anchor=north west,xshift=0.1cm] (image3) at (image2.north east) {\includegraphics[width = 0.3\textwidth,trim={1.5cm 1.5cm 0cm 1.cm},clip]{img/move_mesh_table_4.png}};
	
	\node[label,scale=1., anchor=south west, rectangle, fill=white, align=center, font=\scriptsize\sffamily] (n_0) at (image1.south west) {$t = 0$};
    \node[label,scale=1., anchor=south west, rectangle, fill=white, align=center, font=\scriptsize\sffamily] (n_1) at (image2.south west) {$t = \SI{12}{\second}$};
    \node[label,scale=1., anchor=south west, rectangle, fill=white, align=center, font=\scriptsize\sffamily] (n_1) at (image3.south west) {$t = \SI{38}{\second}$};
	\end{tikzpicture}
	\vspace{-.5em}
    \caption{Scenario~1: Two persons interacting with five chairs and a table, colored by instance ID, in our cluttered lab environment. Four chairs and the table are being moved. The purple chair is standing still.} 
        \label{fig:experiment1}
\end{figure*}
\begin{table}[t]
\caption{IoU scores  for scenarios with ground truth segmentation and PnP + ICP refinement on sensor boards.}
\vspace{-.5em}
\label{tab:gt-segmentation-iou}
\centering
\setlength{\tabcolsep}{2.25pt}
\begin{threeparttable}
\begin{tabular}{c|c|cc|cc|cc|cc|c}
  \toprule
  &  & \multicolumn{2}{c|}{Cam 1} & \multicolumn{2}{c|}{Cam 2} & \multicolumn{2}{c|}{Cam 3} & \multicolumn{2}{c|}{Cam 4} & Total\\\midrule
  Scenario & Type & $E_\text{IoU}$ & $\sigma$ & $E_\text{IoU}$ & $\sigma$ & $E_\text{IoU}$ & $\sigma$ & $E_\text{IoU}$ & $\sigma$ & $E_\text{IoU}$ \\\midrule
  \multirow{2}{*}{\shortstack{\textit{chairblack}\\\textit{hand}}}  &  Mesh  &  \textbf{0.77} & 0.08 & \textbf{0.81} & 0.08 & \textbf{0.78} & 0.08 & \textbf{0.57} & 0.08 & \textbf{0.73} \\
   &  Submap  &  0.61 & 0.07 & 0.68 & 0.09 & 0.57 & 0.09 & 0.54 & 0.08 & 0.60 \\\midrule
  
  \multirow{2}{*}{\shortstack{\textit{chairblack}\\\textit{sit}}}  &  Mesh  &  0.39 & 0.16 & \textbf{0.55} & 0.19 & \textbf{0.50} & 0.13 & 0.57 & 0.13 & \textbf{0.50} \\
   &  Submap  &  \textbf{0.42} & 0.12 & 0.33 & 0.16 & 0.46 & 0.10 & \textbf{0.65} & 0.09 & 0.47 \\\midrule
  
  \multirow{2}{*}{\shortstack{\textit{chairwood}\\\textit{hand}}}  &  Mesh  &  0.62 & 0.1 & \textbf{0.65} & 0.09 & 0.62 & 0.07 & \textbf{0.69} & 0.10 & 0.61 \\
   &  Submap  &  \textbf{0.66} & 0.11 & 0.63 & 0.14 & \textbf{0.68} & 0.12 & 0.67 & 0.12 & \textbf{0.66} \\\midrule
  
  \multirow{2}{*}{\shortstack{\textit{chairwood}\\\textit{sit}}}  &  Mesh  &  0.57 & 0.13 & \textbf{0.56} & 0.07 & 0.57 & 0.09 & \textbf{0.64} & 0.16 & \textbf{0.59} \\
   &  Submap  &  \textbf{0.61} & 0.10 & \textbf{0.56} & 0.11 & \textbf{0.61} & 0.08 & 0.54 & 0.11 & 0.58 \\\midrule
  
  \multirow{2}{*}{\shortstack{\textit{tablesquare}\\\textit{move}}}  &  Mesh  &  \textbf{0.75} & 0.13  &  \textbf{0.75} & 0.11  &  0.69 & 0.11  &  0.53   & 0.09  &  0.68   \\
   &  Submap  &  0.72 & 0.14  &  0.74 & 0.15  &  \textbf{0.78} & 0.15  &  \textbf{0.65}   & 0.15  &  \textbf{0.72} \\
  \bottomrule
\end{tabular}
\end{threeparttable}
\end{table}
\begin{table}[t]
\caption{IoU scores for scenarios with online segmentation and detection and PnP + ICP refinement on sensor boards.}
\vspace{-.5em}
\label{tab:online-segmentation-iou}
\centering
\setlength{\tabcolsep}{2.25pt}
\begin{threeparttable}
\begin{tabular}{c|c|cc|cc|cc|cc|c}
  \toprule
  &  & \multicolumn{2}{c|}{Cam 1} & \multicolumn{2}{c|}{Cam 2} & \multicolumn{2}{c|}{Cam 3} & \multicolumn{2}{c|}{Cam 4} & Total\\\midrule
  Scenario & Type & $E_\text{IoU}$ & $\sigma$ & $E_\text{IoU}$ & $\sigma$ & $E_\text{IoU}$ & $\sigma$ & $E_\text{IoU}$ & $\sigma$ & $E_\text{IoU}$ \\\midrule
  \multirow{2}{*}{\shortstack{\textit{chairblack}\\\textit{hand}}}  &  Mesh  & \textbf{0.70} & 0.10 & \textbf{0.75} & 0.09 & \textbf{0.68} & 0.11 & \textbf{0.56} & 0.11 & \textbf{0.67}\\
   &  Submap  & 0.55 & 0.10 & 0.64 & 0.08 & 0.54 & 0.10 & \textbf{0.56} & 0.09 & 0.57 \\\midrule
  
  \multirow{2}{*}{\shortstack{\textit{chairblack}\\\textit{sit}}}  &  Mesh  & \textbf{0.48} & 0.26 & \textbf{0.50} & 0.26 & \textbf{0.53} & 0.27 & 0.52 & 0.16 & \textbf{0.51}\\
   &  Submap  & 0.43 & 0.13 & 0.32 & 0.16 & 0.41 & 0.16 & \textbf{0.61} & 0.11 & 0.44  \\\midrule
  
  \multirow{2}{*}{\shortstack{\textit{chairwood}\\\textit{hand}}}  &  Mesh  &  \textbf{0.56} & 0.10 & \textbf{0.57} & 0.11 & \textbf{0.61} & 0.09 & \textbf{0.65} & 0.12 & \textbf{0.59} \\
   &  Submap  & 0.51 & 0.09 & 0.49 & 0.10 & 0.55 & 0.08 & 0.58 & 0.09 & 0.53 \\\midrule
  
  \multirow{2}{*}{\shortstack{\textit{chairwood}\\\textit{sit}}}  &  Mesh  &  \textbf{0.56} & 0.14 & \textbf{0.61} & 0.08 & \textbf{0.55} & 0.10 & \textbf{0.60} & 0.12 & \textbf{0.58} \\
   &  Submap  &  0.43 & 0.16 & 0.43 & 0.10 & 0.31 & 0.11 & 0.33 & 0.14 & 0.38 \\\midrule
  
  \multirow{2}{*}{\shortstack{\textit{tablesquare}\\\textit{move}}}  &  Mesh  &   \textbf{0.68} & 0.13  &  \textbf{0.68} & 0.12  &  0.67 & 0.09  &  0.46   & 0.11  &  \textbf{0.62} \\
   &  Submap  &  0.60 & 0.10  &  0.61 & 0.11  &  \textbf{0.69} & 0.10  &  \textbf{0.58}   & 0.14  &  \textbf{0.62} \\
  \bottomrule
\end{tabular}
\end{threeparttable}
\end{table}

We further evaluate the pose and geometry estimation accuracy by calculating the Intersection over Union (IoU) between the estimated object models reprojected into the individual camera views and the respective ground-truth segmentation mask from the dataset annotations.
The reprojection of the 3D object model into the camera views is illustrated in \reffig{fig:backprojection-mesh} and \reffig{fig:backprojection-submap} for the object model mesh and volumetric sub-map representations, respectively.
As the 3D object mesh originates from an offline 3D scan of the object, it represents fine structures, such as armrests or legs in high detail. The fine structures of the chair align well with the images of all four camera perspectives in \reffig{fig:backprojection-mesh}, showing high accuracy of the estimated object pose.
The sub-map, on the other hand, has a discrete spatial resolution and cannot accurately represent the chair legs. Furthermore, it is affected by noisy point cloud measurements.

\reftab{tab:gt-segmentation-iou} and \reftab{tab:online-segmentation-iou} report quantitative results of the IoU evaluation, using ground-truth and online point cloud segments and object detections as input, respectively, and compare the mesh and sub-map object representations.
For a fair comparison between mesh and sub-map, the annotated image segmentation masks are extended with a $10\times10$ dilation kernel to match the discrete \SI{5}{\centi\meter} spatial resolution of the sub-maps.
With ground-truth point cloud segment inputs, the mesh-based object representation performs better than the sub-maps in three of five scenarios, averaged over the four cameras.
With online segmentation and detection inputs, the mesh-based representation performs better or equal in all five scenarios. The IoU is slightly lower in the real-world scenarios, accounting for the higher noise in the input data.

\subsection{Real-world Experiments}
We further demonstrate the performance of our proposed method in real-world scenarios in a highly-cluttered, dynamic environment in three different scenarios.
Here, we employ offline 3D scans of the office chairs and table present in the environment as object models.
Object pose estimation and geometry update are calculated online, in real time.

\paragraph*{Scene Perception in three Scenarios}
In Scenario~1, two persons interact with five chairs and a table in our cluttered lab environment, as shown in \reffig{fig:experiment1}. Point measurements of the tracked objects are not included in the allocentric map of the static geometry. The chairs and table are represented by their respective prior 3D mesh model transformed to the estimated pose. The movement of the objects through the scene is tracked online, in real time.

\begin{figure}[t]
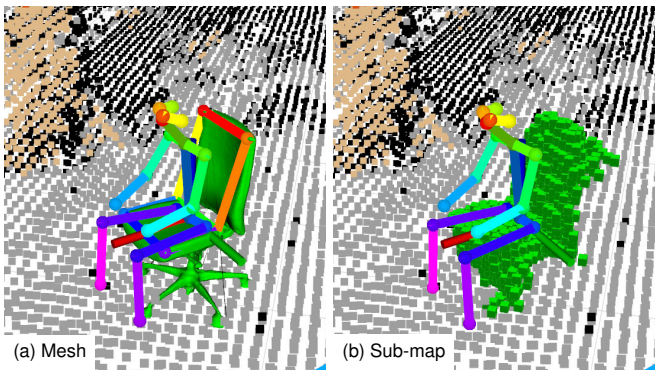

	\centering
	\begin{tikzpicture}
	\node[inner sep=0,anchor=north west] (image1) at (0, 0){\includegraphics[width = 0.235\textwidth,trim={8cm 8cm 9cm 4cm},clip]{img/occluded_mesh.png}};
	\node[inner sep=0,anchor=north west,xshift=0.1cm] (image2) at (image1.north east) {\includegraphics[width = 0.235\textwidth,trim={8cm 8cm 9cm 4cm},clip]{img/occluded_submap.png}};
	
	\node[label,scale=1., anchor=south west, rectangle, fill=white, align=center, font=\scriptsize\sffamily] (n_0) at (image1.south west) {(a) Mesh};
    \node[label,scale=1., anchor=south west, rectangle, fill=white, align=center, font=\scriptsize\sffamily] (n_1) at (image2.south west) {(b) Sub-map};
	\end{tikzpicture}
	\vspace{-.5em}
	\caption{Scenario~2: A chair being occluded by a sitting person. The pose estimate and geometry representation remain stable under high occlusion and interaction between person and object is explained in a physically plausible manner by the scene model.}
    \label{fig:experiment2}
\end{figure}

In Scenario~2 (Fig.~\ref{fig:experiment2}), a person is sitting on a chair, occluding it partially. The estimated object models remain stable also under high occlusion and interactions between persons and objects are explained in a physically plausible manner.
\begin{figure}[t]
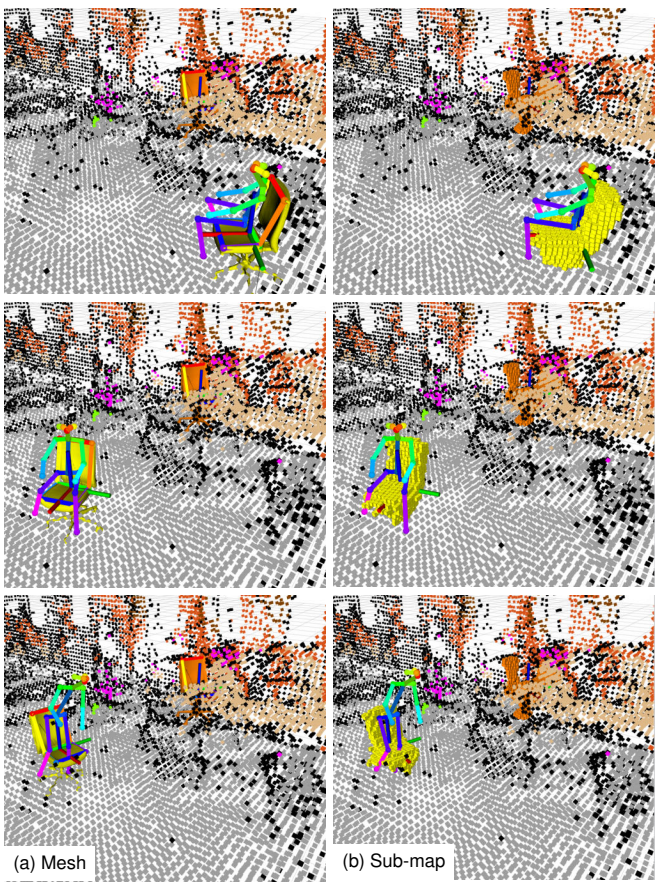

	\centering
	\begin{tikzpicture}
	\node[inner sep=0,anchor=north west] (image1) at (0, 0){\includegraphics[width = 0.235\textwidth,trim={2cm 1cm 0cm 1cm},clip]{img/occluded_move_mesh_sync_1.png}};
	\node[inner sep=0,anchor=north west,xshift=0.1cm] (image2) at (image1.north east) {\includegraphics[width = 0.235\textwidth,trim={2cm 1cm 0cm 1cm},clip]{img/occluded_move_submap_sync_1.png}};
	\node[inner sep=0,anchor=north west,yshift=-0.1cm] (image3) at (image1.south west){\includegraphics[width = 0.235\textwidth,trim={2cm 1cm 0cm 1cm},clip]{img/occluded_move_mesh_sync_2.png}};
	\node[inner sep=0,anchor=north west,xshift=0.1cm] (image4) at (image3.north east) {\includegraphics[width = 0.235\textwidth,trim={2cm 1cm 0cm 1cm},clip]{img/occluded_move_submap_sync_2.png}};
	\node[inner sep=0,anchor=north west,yshift=-0.1cm] (image5) at (image3.south west){\includegraphics[width = 0.235\textwidth,trim={2cm 1cm 0cm 1cm},clip]{img/occluded_move_mesh_sync_3.png}};
	\node[inner sep=0,anchor=north west,xshift=0.1cm] (image6) at (image5.north east) {\includegraphics[width = 0.235\textwidth,trim={2cm 1cm 0cm 1cm},clip]{img/occluded_move_submap_sync_3.png}};
	\node[label,scale=1., anchor=south west, rectangle, fill=white, align=center, font=\scriptsize\sffamily] (n_0) at (image5.south west) {(a) Mesh};
    \node[label,scale=1., anchor=south west, rectangle, fill=white, align=center, font=\scriptsize\sffamily] (n_1) at (image6.south west) {(b) Sub-map};
	\end{tikzpicture}
	\vspace{-.5em}
    \caption{Scenario~3: The yellow chair is being moved (from top-row to bottom) while being occluded by a sitting person. The object movement can be tracked consistently despite the occlusion.}
    \vspace{-1.em}
    \label{fig:experiment3}
\end{figure}

In Scenario~3 (Fig.~\ref{fig:experiment3}), a chair is being moved while being occluded by a sitting person. The movement of the object through the scene can be tracked by our proposed approach even under high occlusions.

\paragraph*{Network Bandwidth}
\label{sec:eval_datarate}
We evaluate the required communication bandwidth for the different object representations. Object detection and tracking run at an update rate of \SI{1}{\hertz} in our sensor network.
The semantic object properties, consisting of object pose, point segment variance, and data association score amount to only 84 Bytes per detected object instance, requiring very low network bandwidth.
The object model and keypoint definitions are a-priori known on both backend and sensors and need not be transmitted during online operation.
The required network bandwidth significantly rises when using the sub-map representation, as raw point cloud segments are transmitted from sensors to backend for each detected object. A point cloud segment of an average size of 250 points amounts to $\sim$\SI{9}{\kilo\byte} of transmitted data.

\section{Conclusions}
\label{sec:Conclusion}
In this work, we extended a framework for multi-view allocentric semantic mapping by a smart edge sensor network~\cite{bultmann_ias2022} with object-level information, using different object representations.
Our method enables pose estimation and tracking of dynamic objects through the static scene geometry. Objects thereby are represented via an a-priori known 3D mesh model or a volumetric sub-map that is learned online. Viewpoints of multiple static smart edge sensors are fused on a central backend. 
Only a few semantic object properties, such as estimated pose and point measurement distribution variances are transmitted from the sensors to the backend, requiring little network bandwidth. Only when volumetric sub-maps are required on the backend, the raw point cloud segments associated to object instances are additionally transmitted, significantly increasing the network traffic.

The object pose estimation follows a two-stage approach of keypoint detection and PnP pose estimation. Poses are further refined using associated point cloud segments via ICP alignment. As the keypoint detection CNNs are trained only on synthetic data, the method can easily be extended to different object classes.
We quantitatively evaluate the pose estimation accuracy of our approach on the public Behave dataset, showing pose errors below \SI{9}{\centi\meter} and \SI{9}{\degree} with online input data processing using lightweight CNN architectures efficient on the embedded sensor hardware.
We demonstrate the application of our method in a cluttered real-world lab environment, where multiple chairs and a table are tracked through the scene online, in real time even under high occlusions.

\section*{Acknowledgment}
The authors would like to thank Malte Splietker for creating 3D models of the chairs and table used for the real-world experiments.

\IEEEtriggeratref{16}
\bibliographystyle{IEEEtran}
\bibliography{literature}

\begin{thebibliography}{10}
\providecommand{\url}[1]{#1}
\csname url@samestyle\endcsname
\providecommand{\newblock}{\relax}
\providecommand{\bibinfo}[2]{#2}
\providecommand{\BIBentrySTDinterwordspacing}{\spaceskip=0pt\relax}
\providecommand{\BIBentryALTinterwordstretchfactor}{4}
\providecommand{\BIBentryALTinterwordspacing}{\spaceskip=\fontdimen2\font plus
\BIBentryALTinterwordstretchfactor\fontdimen3\font minus
  \fontdimen4\font\relax}
\providecommand{\BIBforeignlanguage}[2]{{%
\expandafter\ifx\csname l@#1\endcsname\relax
\typeout{** WARNING: IEEEtran.bst: No hyphenation pattern has been}%
\typeout{** loaded for the language `#1'. Using the pattern for}%
\typeout{** the default language instead.}%
\else
\language=\csname l@#1\endcsname
\fi
#2}}
\providecommand{\BIBdecl}{\relax}
\BIBdecl

\bibitem{bultmann_ias2022}
S.~Bultmann and S.~Behnke, ``{3D} semantic scene perception using distributed
  smart edge sensors,'' in \emph{Int. Conf. on Intelligent Autonomous Systems
  (IAS)}, 2022.

\bibitem{zappel20216d}
M.~Zappel, S.~Bultmann, and S.~Behnke, ``{6D} object pose estimation using
  keypoints and part affinity fields,'' in \emph{RoboCup Int. Symposium}, 2021,
  pp. 78--90.

\bibitem{stilleben_2020}
M.~Schwarz and S.~Behnke, ``Stillleben: Realistic scene synthesis for deep
  learning in robotics,'' in \emph{IEEE Int. Conf. on Robotics and Automation
  (ICRA)}, 2020, pp. 10\,502--10\,508.

\bibitem{sl-cutscenes}
\BIBentryALTinterwordspacing
A.~Boltres, A.~Villar-Corrales, J.~Nogga, and P.~Sch{\"u}tt, ``Sl-cutscenes,''
  2022. [Online]. Available: \url{https://github.com/AIS-Bonn/sl-scenes}
\BIBentrySTDinterwordspacing

\bibitem{EPnP}
V.~Lepetit, F.~{Moreno-Noguer}, and P.~Fua, ``\BIBforeignlanguage{en}{{EPnP}:
  An accurate {O}(n) solution to the {PnP} problem},''
  \emph{\BIBforeignlanguage{en}{International Journal of Computer Vision}},
  vol.~81, no.~2, p. 155, 2008.

\bibitem{behave_2022}
B.~L. Bhatnagar, X.~Xie, I.~Petrov, C.~Sminchisescu, C.~Theobalt, and
  G.~Pons-Moll, ``Behave: Dataset and method for tracking human object
  interactions,'' in \emph{{IEEE} Conference on Computer Vision and Pattern
  Recognition (CVPR)}, 2022.

\bibitem{Bultmann_RSS_2021}
S.~Bultmann and S.~Behnke, ``Real-time multi-view {3D} human pose estimation
  using semantic feedback to smart edge sensors,'' in \emph{Robotics: Science
  and Systems (RSS)}, 2021.

\bibitem{hornung_octomap_2013}
A.~Hornung, K.~M. Wurm, M.~Bennewitz, C.~Stachniss, and W.~Burgard,
  ``{OctoMap}: An efficient probabilistic {3D} mapping framework based on
  octrees,'' \emph{Autonomous Robots}, 2013.

\bibitem{voxblox}
H.~Oleynikova, Z.~Taylor, M.~Fehr, R.~Siegwart, and J.~Nieto, ``Voxblox:
  Incremental {3D} euclidean signed distance fields for on-board {MAV}
  planning,'' in \emph{IEEE/RSJ Int. Conf. on Intelligent Robots and Systems
  (IROS)}, 9 2017.

\bibitem{stuckler_semantic_2012}
J.~Stuckler, N.~Biresev, and S.~Behnke, ``Semantic mapping using object-class
  segmentation of {RGB}-{D} images,'' in \emph{IEEE/RSJ Int. Conf. on
  Intelligent Robots and Systems (IROS)}, 2012, pp. 3005--3010.

\bibitem{maskFusion}
M.~R{\"u}nz, M.~Buffier, and L.~Agapito, ``{MaskFusion}: Real-time recognition,
  tracking and reconstruction of multiple moving objects,'' in \emph{IEEE
  International Symposium on Mixed and Augmented Reality (ISMAR)}, 2018, pp.
  10--20.

\bibitem{he_maskrcnn_2017}
K.~He, G.~Gkioxari, P.~Dollár, and R.~Girshick, ``{Mask R-CNN},'' in
  \emph{IEEE Int. Conf. on Computer Vision (ICCV)}, 2017, pp. 2980--2988.

\bibitem{mid}
B.~Xu, W.~Li, D.~Tzoumanikas, M.~Bloesch, A.~Davison, and S.~Leutenegger,
  ``{MID}-fusion: Octree-based object-level multi-instance dynamic {SLAM},'' in
  \emph{IEEE Int. Conf. on Robotics and Automation (ICRA)}, 2019, pp.
  5231--5237.

\bibitem{voxblox++}
M.~Grinvald, F.~Furrer, T.~Novkovic, J.~J. Chung, C.~Cadena, R.~Siegwart, and
  J.~Nieto, ``Volumetric instance-aware semantic mapping and {3D} object
  discovery,'' \emph{IEEE Robotics and Automation Letters}, vol.~4, no.~3, pp.
  3037--3044, 2019.

\bibitem{tsdf++}
M.~Grinvald, F.~Tombari, R.~Siegwart, and J.~Nieto, ``{TSDF++}: A multi-object
  formulation for dynamic object tracking and reconstruction,'' in \emph{IEEE
  Int. Conf. on Robotics and Automation (ICRA)}, 2021, pp. 14\,192--14\,198.

\bibitem{cao_openpose_2018}
Z.~Cao, G.~Hidalgo, T.~Simon, S.-E. Wei, and Y.~Sheikh, ``{OpenPose}: Realtime
  multi-person {2D} pose estimation using part affinity fields,'' \emph{IEEE
  Transactions on Pattern Analysis and Machine Intelligence}, vol.~43, no.~1,
  pp. 172--186, 2021.

\bibitem{xiao_simple_2018}
B.~Xiao, H.~Wu, and Y.~Wei, ``Simple baselines for human pose estimation and
  tracking,'' in \emph{Europ. Conf. on Computer Vision (ECCV)}, 2018, pp.
  466--481.

\bibitem{PoseCNN}
Y.~Xiang, T.~Schmidt, V.~Narayanan, and D.~Fox, ``{PoseCNN}: A convolutional
  neural network for {6D} object pose estimation in cluttered scenes,'' in
  \emph{Robotics: Science and Systems (RSS)}, 2018.

\bibitem{Ransac}
M.~A. Fischler and R.~C. Bolles, ``Random sample consensus: A paradigm for
  model fitting with applications to image analysis and automated
  cartography,'' \emph{Communications of the ACM}, vol.~24, no.~6, pp.
  381--395, 1981.

\bibitem{mobilenetv32019}
A.~Howard, M.~Sandler, G.~Chu, L.-C. Chen, B.~Chen, M.~Tan, W.~Wang, Y.~Zhu,
  R.~Pang, V.~Vasudevan, Q.~V. Le, and H.~Adam, ``{Searching for
  MobileNetV3},'' in \emph{IEEE Int. Conf. on Computer Vision (ICCV)}, 2019,
  pp. 1314--1324.

\bibitem{paetzold_camcalib_2022}
B.~P{\"a}tzold, S.~Bultmann, and S.~Behnke, ``Online marker-free extrinsic
  camera calibration using person keypoint detections,'' in \emph{DAGM German
  Conference on Pattern Recognition (GCPR)}, 2022.

\bibitem{pcl_icra_2011}
R.~B. Rusu and S.~Cousins, ``{3D} is here: Point cloud library ({PCL}),'' in
  \emph{IEEE Int. Conf. on Robotics and Automation (ICRA)}, 2011.

\bibitem{xiong_mobiledets_2021}
Y.~Xiong, H.~Liu, S.~Gupta, B.~Akin, G.~Bender, Y.~Wang, P.-J. Kindermans,
  M.~Tan, V.~Singh, and B.~Chen, ``{MobileDets}: Searching for object detection
  architectures for mobile accelerators,'' in \emph{IEEE Conf. on Computer
  Vision and Pattern Recognition (CVPR)}, 2021, pp. 3825--3834.

\bibitem{lin_coco_2014}
T.-Y. Lin, M.~Maire, S.~Belongie, J.~Hays, P.~Perona, D.~Ramanan,
  P.~Doll{\'a}r, and C.~L. Zitnick, ``{Microsoft COCO: Common objects in
  context},'' in \emph{Europ. Conf. on Computer Vision (ECCV)}, 2014, pp.
  740--755.

\end{thebibliography}

\end{document}